\documentclass{article} % For LaTeX2e
\usepackage{iclr2021_conference,times}

\usepackage{amsmath,amsfonts,bm}

\def\eqref#1{equation~\ref{#1}}
\def\1{\bm{1}}

\DeclareMathAlphabet{\mathsfit}{\encodingdefault}{\sfdefault}{m}{sl}
\SetMathAlphabet{\mathsfit}{bold}{\encodingdefault}{\sfdefault}{bx}{n}

\usepackage{graphicx}
\usepackage{hyperref}
\usepackage{url}

\title{Fourier Transform Approximation as an Auxiliary Task for Image Classification}

\author{Chen Liu\\
\texttt{cl3760@columbia.edu}
}

\begin{document}
\maketitle

\begin{abstract}
Image reconstruction is likely the most predominant auxiliary task for image classification, but we would like to think twice about this convention. In this paper, we investigated ``approximating the Fourier Transform of the input image" as a potential alternative, in the hope that it may further boost the performances on the primary task or introduce novel constraints not well covered by image reconstruction. We experimented with five popular classification architectures on the CIFAR-10 dataset, and the empirical results indicated that our proposed auxiliary task generally improves the classification accuracy. More notably, the results showed that in certain cases our proposed auxiliary task may enhance the classifiers' resistance to adversarial attacks generated using the fast gradient sign method.
\end{abstract}

\section{Introduction}
Auxiliary learning~\citep{Auxiliary1, Auxiliary2}, a smart regularization approach, is often used to improve the performances of a deep learning model by preventing overfitting and enhancing its ability to generalize on unseen data. This is achieved by introducing an "auxiliary task" in parallel to the actual purpose of the model, i.e. the "primary task (or primary tasks)", and incorporating its effects as an additional factor in the loss function. In general, the auxiliary task does not semantically overlap with the primary task, but it is usually designed such that enforcing good performances on the auxiliary task shall naturally and intuitively boost the model's capabilities on the primary task.

In image classification, ``reconstructing the input image" is probably the most popular auxiliary task~\citep{Recon1, Recon2, Recon3}. For simplicity, we will use \textit{Recon} to denote this auxiliary task. Researchers claimed that doing so helps the neural networks extract meaningful high-level features during the encoding path (the sequences of convolution, activation, pooling, etc) where image features are encoded into the feature space. By ensuring that these feature representations can be decoded back into near replicas of the original input image, the aforementioned \textit{Recon} auxiliary task is able to, at least to some extent, supervise the model to not encode features in a nonsensical manner. One particular elegance in this task is that no additional human labor is required -- all what we need are already provided by default -- it's the input image itself.

While \textit{Recon} has great merits as an auxiliary task for image classification, it is natural for us to further explore other alternatives. Does there exist another auxiliary task that imposes a stronger regularization and thus is more helpful to the primary task? Or can we find an alternative that poses constraints on some aspect that is weakly constrained by \textit{Recon}?

Going along these lines, in this paper we propose ``approximating the Fourier Transform of the input image" as an alternative auxiliary task. For simplicity, we will use \textit{FT} to denote this auxiliary task. Similar to the \textit{Recon} task, \textit{FT} does not require any additional labeling since the ground truth Fourier Transform can be directly computed from the input image. Moreover, it has the potential to enforce the model to encode robust features in the frequency domain. This might be a beneficial property under certain context, such as in improving the model's robustness against adversarial attacks, as some recent works~\citep{AdversarialFreq1, AdversarialFreq2} have suggested the possibility that adversarial attacks may be more recognizable in the frequency domain than in the spatial domain.

\section{Methods}
\subsection{Dataset}
We used the CIFAR-10 dataset~\citep{CIFAR-10}, a popular image classification dataset with 6,000 RGB images of dimension (32$\times$32$\times$3) in each of the 10 classes. Unlike most papers using this dataset that strictly comply with the original train-test split, we decided to further split the original train set into train and validation sets~(95\%:5\%), due to considerations similar to what was stated in \citet{Faster_Adv}. As compared to training on the 50,000 train images and assessing the model performances on the 10,000 test images at each epoch (as in most previous works), we kept the test images stand-alone until the final evaluation, and used the performances on the validation set to gauge learning rate scheduling and early stopping. In this way, we reduced the risk of overfitting and deflated the test set performances.

Images in the train, validation and test sets all underwent intensity standardization process using the mean and standard deviation from the original train set. Beyond that, a minor data augmentation was applied to images in the train and validation sets using random cropping and flipping. The data loaders sample from the train set in random order to reduce the risk of ``the model learning the order rather than the content", while they sample from the validation and test set in sequential order to ensure consistency across different experiments.

\begin{figure}[hb]
\begin{center}
\includegraphics[width=380pt,trim=10 10 10 10,clip]{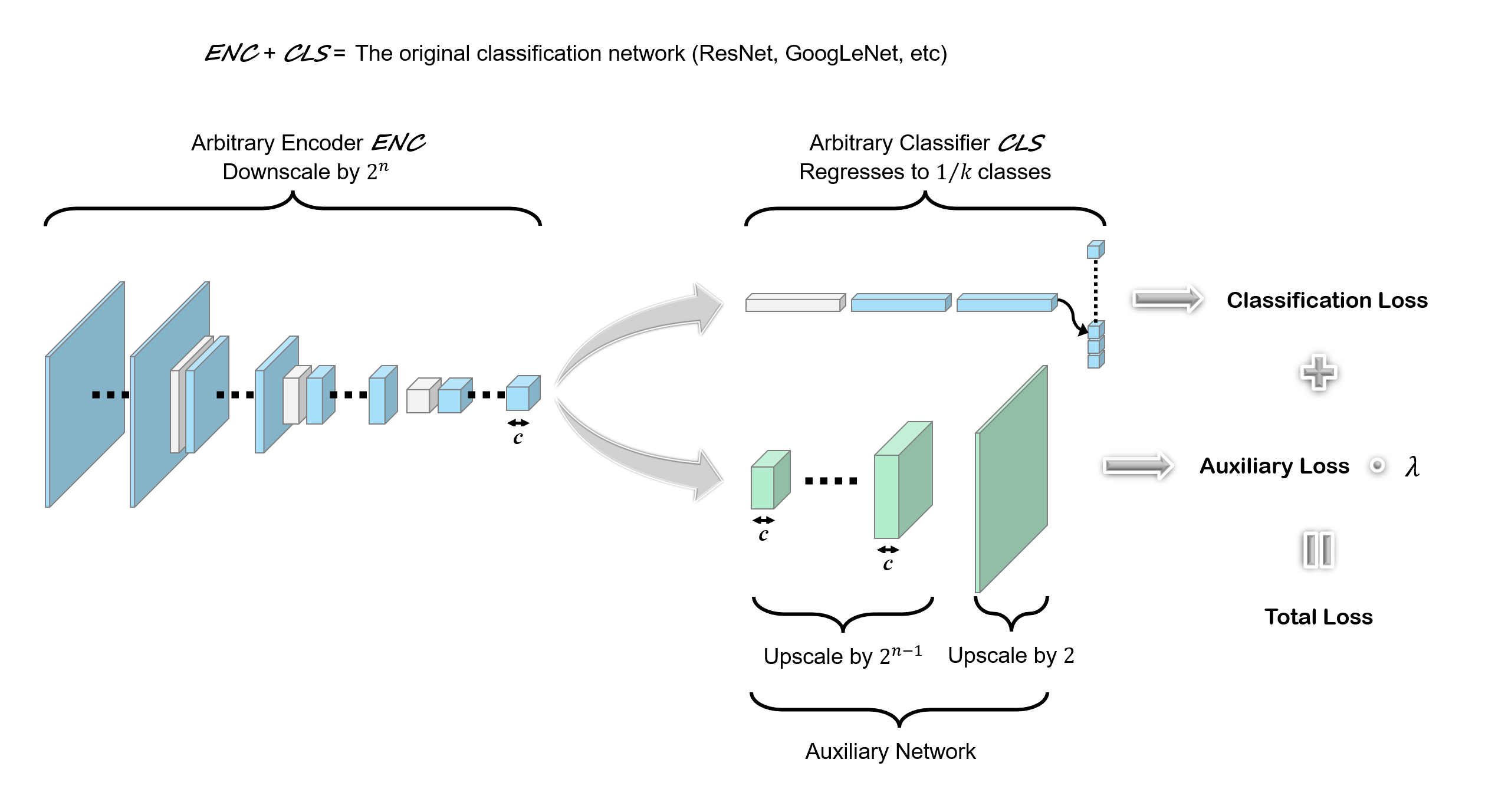}
\end{center}
\caption{Workflow of the auxiliary training process. Our proposed auxiliary network behaves like an decoder that maps the encoded features back to the image space. Depending on the auxiliary task, the output dimension may either be (32$\times$32$\times$3) for \textit{Recon} or (32$\times$32$\times$2) for \textit{FT}. In our implementation, each green block is a PyTorch module \textit{torch.nn.ConvTranspose2d} followed by ReLU, but of course the auxiliary network can be modified or re-designed.}
\label{Workflow}
\end{figure}

\subsection{Model Architectures}
We adapted from an open-source repository\footnote{\url{https://github.com/kuangliu/pytorch-cifar}} which implemented several modern classification architectures, and our code is publicly available on GitHub\footnote{\url{https://github.com/ChenRaphaelLiu/Classification_Auxiliary_Task}}. We selected ResNet~\citep{ResNet}, ResNeXt~\citep{ResNeXt}, GoogLeNet~\citep{GoogLeNet}, EfficientNet~\citep{Efficientnet}, and MobileNetV2~\citep{MobileNetV2}. We did not choose some other equally popular architectures for various reasons. For example, VGG~\citep{VGG} has too many pooling layers. Even the most light-weight version has 5 pooling layers that shrinks the image-space dimension from 32$\times$32 to 1$\times$1, which makes the auxiliary tasks, i.e. \textit{Recon} and \textit{FT}, not very plausible.

\subsection{Primary and Auxiliary Tasks}
Our proposed workflow is illustrated in Fig~\ref{Workflow}. Clearly, this is a generalizable framework, although in this paper we are only using it in its current form. The primary task is the 10-class image classification on CIFAR-10, supervised by a cross entropy loss. The auxiliary tasks are one of the following:

\begin{itemize}
    \item \textit{None} -- The whole network is effectively the same as the original classification network alone. The auxiliary network is not visited in either the forward or backward propagation, and the auxiliary loss is not included in the total loss.
    \item \textit{Recon} -- The auxiliary network aims to reconstruct the input image, and a pixel-level mean squared error is used as the auxiliary loss.
    \item \textit{FT} -- The auxiliary network aims to approximate the Fourier Transform of the input image, with the same mean squared error auxiliary loss. The ground truth is a two-channel Fourier Transformed version of the gray-scaled input image, with the first channel being the magnitude and the second being the phase.
\end{itemize}

The total loss is a weighted sum between the classification loss and the auxiliary loss, as shown in Fig~\ref{Workflow}. $\lambda$ is set to 0.01.

For autonomy and consistency, we used the same training strategy across all the classification network variants. The initial learning rate is set to 0.1, while an adaptive learning rate scheduler reduces the learning rate by 80\% each time the validation accuracy plateaus. When the validation accuracy stops increasing for 10 consecutive epochs, the early stopping mechanism will be triggered, and only the model weights corresponding to the best validation accuracy will be saved.

\begin{table}[hb]
\caption{CIFAR-10 classification accuracies~(unit: percent) on the test set. Note the accuracies may not be optimal since all experiments are conducted under the same training strategy and hyperparameters.}
\begin{center}
\includegraphics[width=380pt,trim=1 1 1 1,clip]{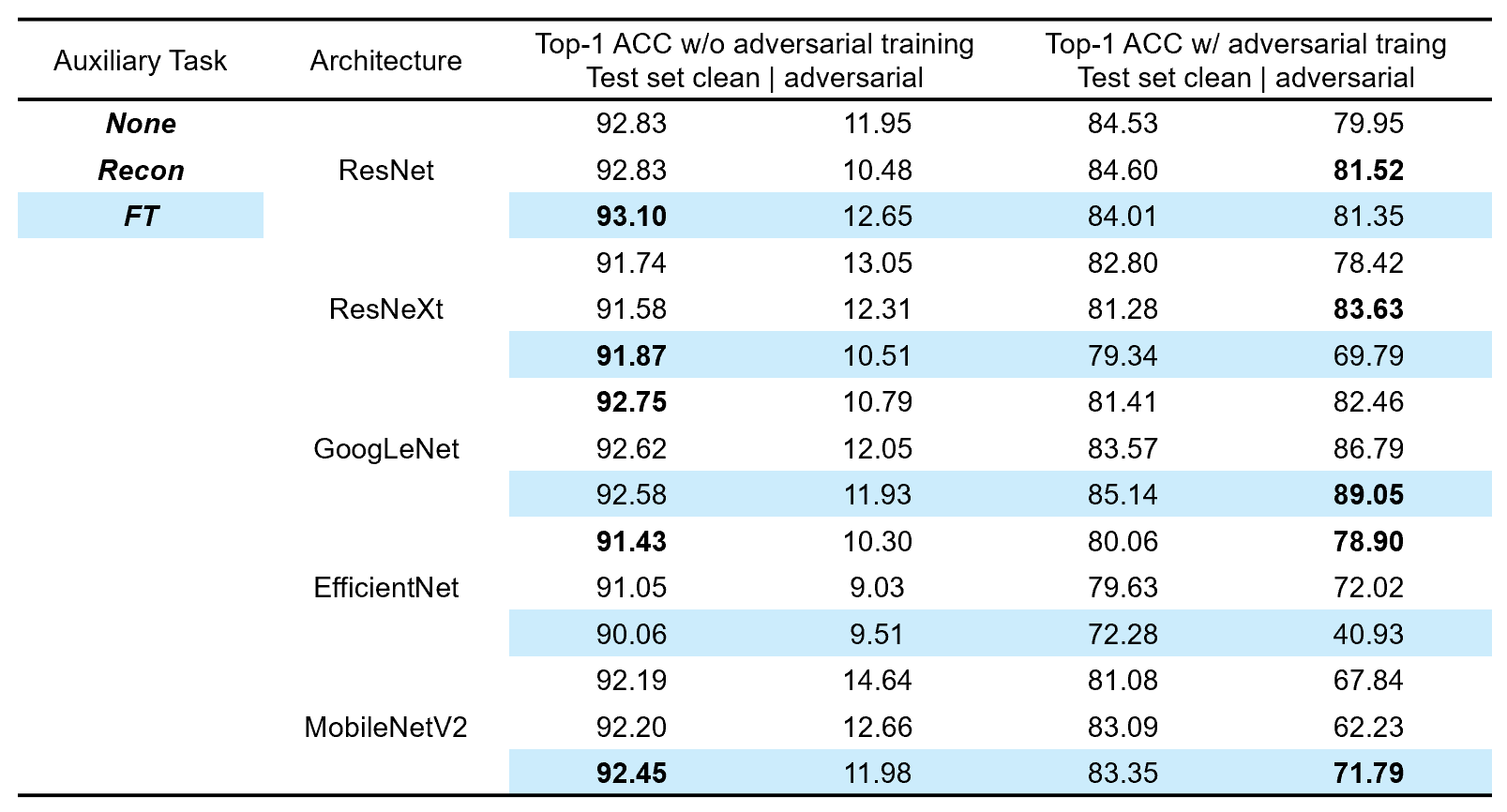}
\end{center}
\label{Accuracies}
\end{table}

\subsection{Adversarial Attack}
Adversarial attacks are generated using the fast gradient sign method~(FGSM)~\citep{FGSM}. We utilized an off-the-shelf library\footnote{\url{https://github.com/cleverhans-lab/cleverhans}} for generating these attacks~\citep{Adversarial_lib} with epsilon at 0.3.

\section{Results}

Classification accuracies are recorded in Table~\ref{Accuracies}. In general, models without adversarial training are very vulnerable to adversarial attacks. Without adversarial training, the top-1 accuracies are usually around 90\% when evaluated on the clean test set but the accuracies catastrophically drop to around 10\% when the test set is adversarially processed. Training with adversarial samples generally degrades the accuracies on the clean test set, but boosts the accuracies on the adversarial test set significantly.

Compared to not using an auxiliary task and using \textit{Recon}, our proposed \textit{FT} auxiliary task resulted in better performances in several cases. For the non-adversarial scenario~(clean training, clean test set), \textit{Recon} achieved the highest top-1 accuracy on 3 out of 5 architectures. For the adversarial scenario~(adversarial training, adversarial test set), \textit{Recon} surpassed the two alternatives on 2 out of 5 architectures.

It worth noting that the EfficientNet backbone does not work well with \textit{Recon} in our experiments. It is exemplified by the very low test accuracy (40.93\%) in the adversarial scenario. We may want to further diagnose the reason behind this outstanding behavior.

\section{Discussions}
This is not meant to be a paper with comprehensive experiments using tons of computational resources. Rather, we are trying to show some preliminary results along a quite random idea. People have been using \textit{Recon} as their go-to auxiliary task and have taken that for granted for quite a long time. We are just trying to provoke some thoughts on whether or not we can find an alternative auxiliary task that has its own merits. For example, \textit{FT} as an auxiliary task might pose constraints during the training process and enforces the encoder to learn robust and meaningful feature in the frequency domain, although we have not done any solid validation to show whether or not it is the case in this paper. Nevertheless, we find it worthwhile to ask the community to re-think the convention of using \textit{Recon} as the ``default" auxiliary task in image classification.

\subsubsection*{Author Contributions}
What do you think? :)

\subsubsection*{Acknowledgments}
We would like to sincerely thank the scalpers who encouraged us to spend almost twice the suggested price on the GPU we used to run the experiments in this paper.

\bibliography{iclr2021_conference}
\bibliographystyle{iclr2021_conference}

\end{document}